\documentclass[letterpaper, 10 pt, journal, twoside]{IEEEtran}
\ifCLASSINFOpdf
\else
\fi

\usepackage{mathpazo,amsmath,amssymb,epsfig,enumerate,bbm,calc,color,ifthen,capt-of} 
\usepackage{algorithm}
\usepackage{color, graphicx}
\usepackage[scaled]{helvet} 
\usepackage{booktabs, multirow}
\usepackage{tabularx}
\usepackage[table]{xcolor}
\usepackage{subcaption}
\usepackage{balance}
\usepackage[justification=raggedright]{caption}

\begin{document}
%
\title{Ground-aware Monocular 3D Object Detection for Autonomous Driving}
%
%
%

\author{Yuxuan Liu$^{1}$, Yuan Yixuan$^{2}$ and Ming Liu$^{1}$%
\thanks{Manuscript received: October, 14, 2020; Revised November, 30, 2020 ; Accepted January, 5, 2021.}
\thanks{This paper was recommended for publication by Youngjin Choi upon evaluation of the Associate Editor and Reviewers' comments.} 
\thanks{$^{1}$Yuxuan Liu and Ming Liu are with the Robotics and Multi-Perception Laborotary, 
Department of Electronic and Computer Engineering, The Hong Kong University of Science and Technology
        {\tt\footnotesize yliuhb@connect.ust.hk ,eelium@ust.hk}}%
\thanks{$^{2} $Yuan Yixuan is with the Department of Electrical Engineering, City University of Hong Kong, Hong Kong, China.
        {\tt\footnotesize yxyuan.ee@cityu.edu.hk}}%
\thanks{Digital Object Identifier (DOI): see top of this page.}
}
%
%

\markboth{IEEE Robotics and Automation Letters. Preprint Version. Accepted January, 2021}
{Yuxuan Liu \MakeLowercase{\textit{et al.}}: Ground-aware Monocular 3D Object Detection for Autonomous Driving} 

%



\maketitle

\begin{abstract}
  Estimating the 3D position and orientation of objects in the environment with a single RGB camera is a critical and challenging task for low-cost urban autonomous driving and mobile robots.
  Most of the existing algorithms are based on the geometric constraints in 2D-3D correspondence, which stems from generic 6D object pose estimation.
  We first identify how the ground plane provides additional clues in depth reasoning in 3D detection in driving scenes.
  Based on this observation, we then improve the processing of 3D anchors and introduce a novel neural network module to fully utilize such application-specific priors in the framework of deep learning.
  Finally, we introduce an efficient neural network embedded with the proposed module for 3D object detection.
  We further verify the power of the proposed module with a neural network designed for monocular depth prediction. The two proposed networks achieve state-of-the-art performances on the KITTI 3D object detection and depth prediction benchmarks, respectively. The code will be published in https://www.github.com/Owen-Liuyuxuan/visualDet3D
\end{abstract}


%
\IEEEpeerreviewmaketitle

\section{Introduction}

Simultaneously estimating the position, orientation, and dimensions of an object in 3D with a single well-calibrated RGB camera image in an autonomous driving scene is generally an ill-posed problem. 
Lidar-based methods and stereo-vision-based methods, which respectively obtain depths and distance information from lidar measurements and triangulation, can achieve superior performance \cite{Yun2018Focal}\cite{wang2020PointTrackNet}\cite{chen2019ImageDe}\cite{Li2019Stereo}. Monocular setups are cheaper and more versatile than LiDAR setups and are more robust to variations in extrinsic parameters than stereo cameras. 
 As a result, 3D detection with a single camera is still a heated research direction despite a lack of depth information.

Recent developments in monocular 3D object detection mainly utilize geometric constraints between the 3D object and the its projection on a 2D image. ShiftRCNN \cite{Li2019ShiftRCNN}, SS3D \cite{Jorgensen2019SS3D}, and RTM3D \cite{Li2020RTM3DRM} optimize the estimation of depth and orientation by solving a Perspective-n-Point problem with noisy observation.

Most of these ideas come from a more general problem of monocular 6D pose estimation. Monocular 6D pose estimation benchmarks like LINEMOD \cite{hinterstoisser2012linemod} are based on the assumption that the CAD models of the objects of interest are known. However, we do not have access to accurate car models for each vehicle in autonomous driving scenes; thus, the performances of monocular 3D object detectors in autonomous driving scenes are limited. 

In autonomous driving and mobile robotics applications, we can generally assume that most important dynamic objects are on a ground plane, and the camera is mounted at a certain height above the ground. Some traditional depth prediction methods also note the importance of ground planes and introduce a similar "floor-wall" assumption for indoor environments \cite{Delage2006FloorWall}, \cite{Chun2013FloorDetectionDepthEstimation}.
Such perspective priors on the ground plane, not presented in \textit{general} monocular 6D pose estimation problems, provide a significant amount of information for geometric reasoning for monocular 3D object detection in driving scenes.  
 Few recent works \textit{explicitly} inject the perspective priors on the ground plane into a neural network.

This paper proposes two novel procedures to allow a monocular object detector to reason over the ground plane explicitly.

The first procedure is anchor filtering, where we explicitly break the invariance in the neural network predictions. Given a prior distance between an anchor and its distance to the camera, we back-project the anchor to 3D. Since all objects of interest are located around the ground plane, we filter out 3D anchors far from the ground plane during training and testing. This operation focuses the network on positions where objects of interest are likely to appear. We will further introduce this procedure in Section~\ref{sec:anchor_method}.

\begin{figure}
  \centering
    \includegraphics[width=0.9\linewidth]{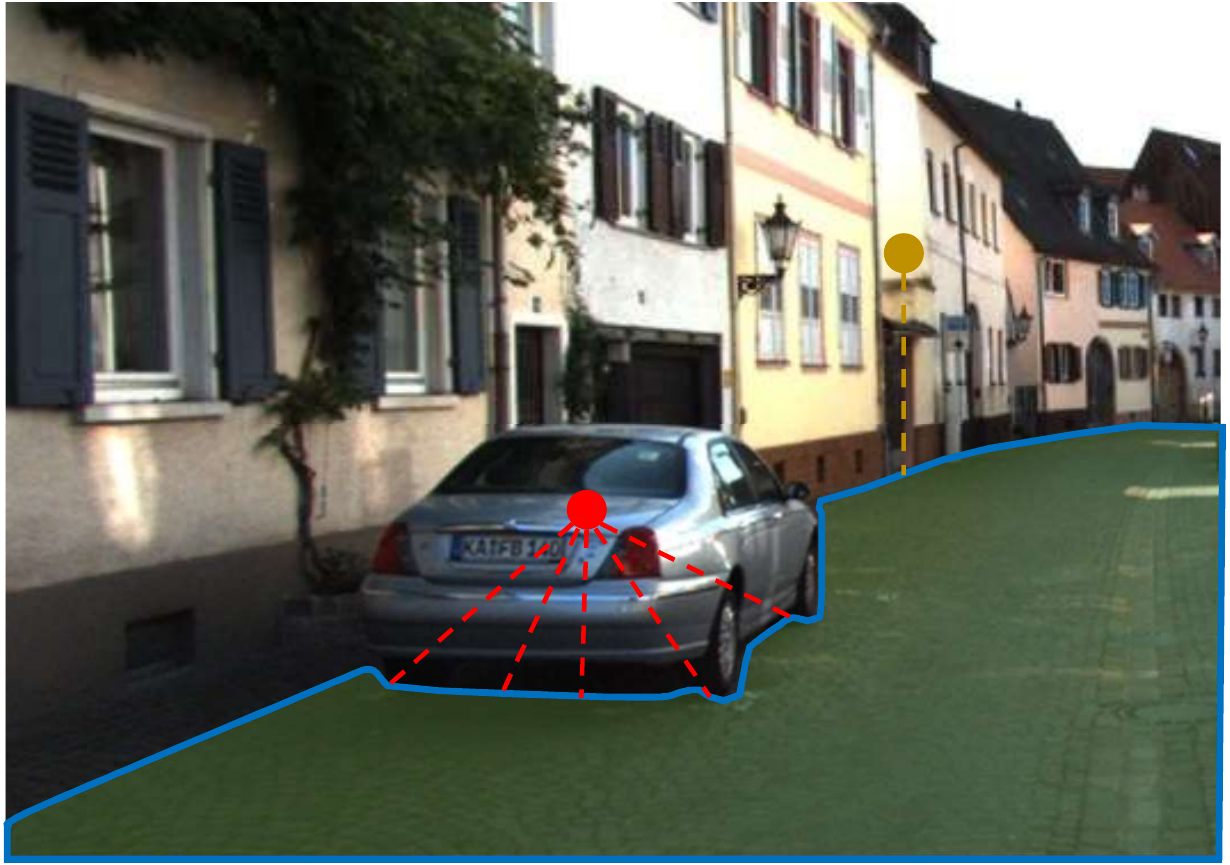}
    \caption{Contact points with the ground plane are important in inferreing 3D information of an object. Predicting depths of background pixels (e.g., the brown point) also rely on the geometry of the ground plane. Best viewed in color.}
    \label{fig:motivation}
  \end{figure}

The second procedure is a ground-aware convolution module. The motivation of this module is illustrated by Figure~\ref{fig:motivation}. For a human, ground pixels around a car are useful to estimate the car's 3D position, orientation, and dimensions. For an anchor-based detector, features at the center are responsible for estimating all the car's 3D parameters.  However, to infer depth with ground pixels like a human, the network model needs to perform the following steps from the center of an object (e.g. the red dot in the figure) \begin{enumerate}
    \item identifing the contact points of the object and the ground plane (e.g. the blue curve beneath the car).
    \item computing the 3D position of the contact points with perspective geometry.
    \item gathering information from these contact points with a receptive field focusing downwards.
\end{enumerate}

A standard object detection or depth prediction network is built to have a uniform receptive field, and neither perspective geometry priors nor camera parameters are provided to the network. Thus, it is non-trivial to train a standard neural network to make inferences like a human.

The ground-aware convolution module is designed to guide the network to incorporate the ground-based reasoning in network inferencing. We encode each pixel point's prior depth value as an additional feature map, and we guide each pixel point of the feature map to incorporate features from pixels below them. Details of this module will be introduced in Section~\ref{sec:lookground}.

Incorporating the two proposed procedures into the network, we propose a one-stage framework with explicit ground plane hypothesis usage. The network is fast thanks to its clean structure and can run at about 20 frames-per-second (FPS) on a modern GPU.

We further incorporate the ground-aware convolution module in a u-net-based structure on monocular depth prediction. Both networks achieve state-of-the-art (SOTA) performance on the KITTI dataset.

The contribution of the paper is three-fold.
\begin{itemize}
    \item We identify the benefit of learning from the ground plane priors in urban scenes for 3D reasoning from images.
    \item We introduce a processing method and a ground-aware convolution module in monocular 3D object detection to use the ground plane hypothesis.
    \item We evaluate the proposed module and design methods on the KITTI 3D object detection benchmark and the depth prediction benchmark, and we achieve competitive results.
\end{itemize}

\section{Related Works}
\label{section:Relate}

\subsection{Pseudo-LiDAR for Monocular 3D Object Detection}
The idea of pseudo-LiDAR, reconstructing point clouds from mono or stereo images, has led to the recent advances in 3D detection \cite{wang2018pseudo}\cite{Ma2019AM3D}\cite{Vianney2019RefinedMPL}\cite{Weng2019Plidar}\cite{Ku2019MonoPSR}.
Pseudo-LiDAR methods usually reconstruct the point cloud from a single RGB image with off-the-shelf depth prediction networks, which limit their performance.
Moreover, the current SOTA monocular depth prediction networks generally take about 0.05s per frame, which significantly limits the inference speed of pseudo-lidar detection pipelines.

\subsection{One-Stage Detection for Monocular 3D Object Detection}
Several recent advances in monocular 3D object detection directly regress 3D bounding boxes in a one-stage object detection framework. 

\textit{Optimization-based Methods:}
SS3D \cite{Jorgensen2019SS3D} concurrently estimated 2D bounding boxes, depth, orientation, dimensions, and 3D corners. Nonlinear optimization was applied to merge all these predictions.
Shift-RCNN\cite{Li2019ShiftRCNN} also estimated 3D information in a 2D anchor and applied a small sub-network instead of a nonlinear solver.
More recent methods, SMOKE\cite{liu2020SMOKE} and RTM3D \cite{Li2020RTM3DRM} incoperate the aforementioned optimization scheme into the anchor-free object detector CenterNet \cite{zhou2019objects}.

\textit{3D Anchor-based Methods:}
M3D-RPN \cite{Brazil2019M3DRPN} introduced 3D priors in 2D anchors, and also emphasized the importance of the ground plane hypothesis.
It also introduced height-wise convolution while D4LCN \cite{Ding2019D4LCN} introduced depth-guided convolution.
Both techniques came at a high cost to efficiency and only utilized the ground plane hypothesis implicitly.

We point out that anchor-based methods are still better than anchor-free methods in 3D detection. Anchor-free detectors implicitly require the network to learn the correlation between the object's apparent size and its distance value. In contrast, anchor-based detectors can embed this in an anchor's preprocessing. As a result, we develop our framework upon anchor-based detectors.

To our knowledge, our proposed framework is the first 3D anchor-based method to explicitly utilize the ground plane hypothesis of driving scenes in monocular 3D detection and achieves the SOTA performance at the time of writing.

\subsection{Supervised Monocular Depth Prediction With Deep Learning}

Supervised monocular depth prediction is another hot research topic closely related to monocular 3D object detection. 

DORN \cite{Fu2018DORN} and SoftDorn \cite{diaz2019SoftDorn} proposed to treat the depth estimation problem as an ordinal regression problem to improve the convergence rate. BTS \cite{Lee2019BTS} proposed the local planar guidance module and incorporated normal information to constraint the depth prediction results in the scenes. BANet \cite{Aich2020BANet}, meanwhile, proposed a bidirectional attention network to improve the receptive fields and global information understanding of depth prediction networks. 

Many of the methods above focus on depth prediction for multiple datasets and scenarios. Images in datasets like NYUv2\cite{SilbermanECCV12NYU} and DIODE\cite{diode_dataset} are taken from various viewpoints, and it is hard to extract floor priors, unlike the cases in driving scenes. As a result, the neural networks mentioned above do not utilize the camera's extrinsic parameters to extract environment priors, and the absolute scale is lacking during the network inference process.

\section{Methods}
\label{section:Methods}

In this section, we elaborate on the methods applied in this paper. First, we present the formulation of the detection network's inference results and the data preprocessing procedure. 
Second, we introduce the ground-aware convolution module that extracts depth priors from the ground plane hypothesis.
Finally, we present the network's architecture with other major modifications in the training and inferencing process.

\subsection{Anchors Preprocessing}
\label{sec:anchor_method}
\subsubsection{Anchors Definition}

We follow the idea from YOLO \cite{yolov3} to densely predict bounding boxes with dense anchors.
Each anchor on the image also acts as a proposal of an object in 3D. A 3D anchor consists of a 2D bounding box
parameterized by $[x, y, w_{2d}, h_{2d}]$, where $(x, y)$ is the center of the 2D box and $(w_{2d}, h_{2d})$ is the width and height;
3D centers of an object are presented as $[cx, cy, z]$, where $(cx, cy)$ is the center of the object projected on the
image plane and $z$ is the depth; $[w_{3d}, h_{3d}, l_{3d}]$ corresponds to the width, height and length of the 3D bounding box, and $[sin(\alpha), cos(\alpha)]$ is the sine and cosine value of the observation angle $\alpha$.

\begin{figure}
    \centering
        \includegraphics[width=1.0\linewidth]{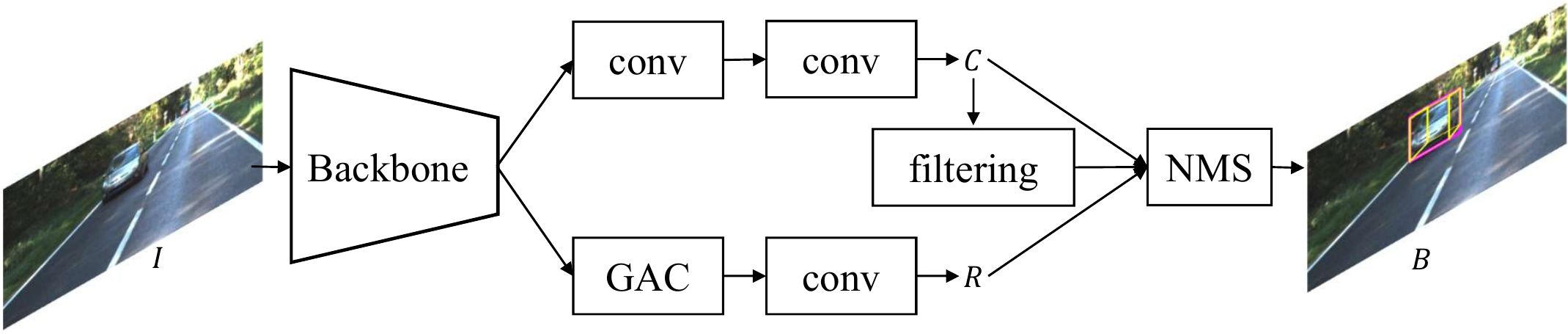}

    \caption{Network structure for 3D object detection. We extract features from image $I$ and predict classfication tensor $C$ and regression tensor $R$. We filter anchors far from the ground before post-processing and produce the final bounding boxes $B$. 
    }
    \label{fig:network}
\end{figure}

\subsubsection{Priors Extraction from Anchors}
The shape and size of an anchor or an object are highly correlated with the depth. In some prior methods \cite{Brazil2019M3DRPN}, the mean of the depth is computed for each pre-defined anchor box, while variance is computed globally instead. The global variance is only computed to normalize the targets for the neural net.

We further observe that the variance of the depth $z$ of an anchor is inversely proportional to the object's size in the image. Thus, we consider each anchor as a distribution with individual mean and variance of the object proposal in 3D. To collect prior statistical knowledge in the anchors, we iterate through the training set and collect all objects sharing a large intersection-over-union (IoU) with the box for each anchor box with a different shape.
Then we calculate the mean and variance of the depth $z$, $sin(\alpha)$ and $cos(\alpha)$ for each pre-defined anchor box. 
We can significantly lower the prior variance of the depth $z$ for large anchor boxes / close objects.

Since we have considered anchors as distribution of 3D proposals, the associated 3D targets should not deviate much from the expectation. We utilize the fact that most objects of interest should be on the ground plane. Each anchor, centering at $(u, v)$ with pre-computed mean depth $\hat z$, can be back-projected to 3D:
\begin{equation}
    x_{3d} = \frac{u - c_x}{f_x} \hat z \;\;\;\;\;\;\;\;\; y_{3d} = \frac{v - c_y}{f_y} \hat z,
\end{equation}
where $(c_x, c_y)$ is the camera's principal point and $(f_x, f_y)$ is the camera's focal length. Anchors with $y_{3d}$ too far from the ground will be filtered out from training and testing. Such a strategy allows the network to train with 3D anchors around the region of interest and simplify the classification problem.

\subsection{Ground-Aware Convolution Module}
\label{sec:lookground}
\begin{figure}
    \centering
        \includegraphics[width=1.0\linewidth]{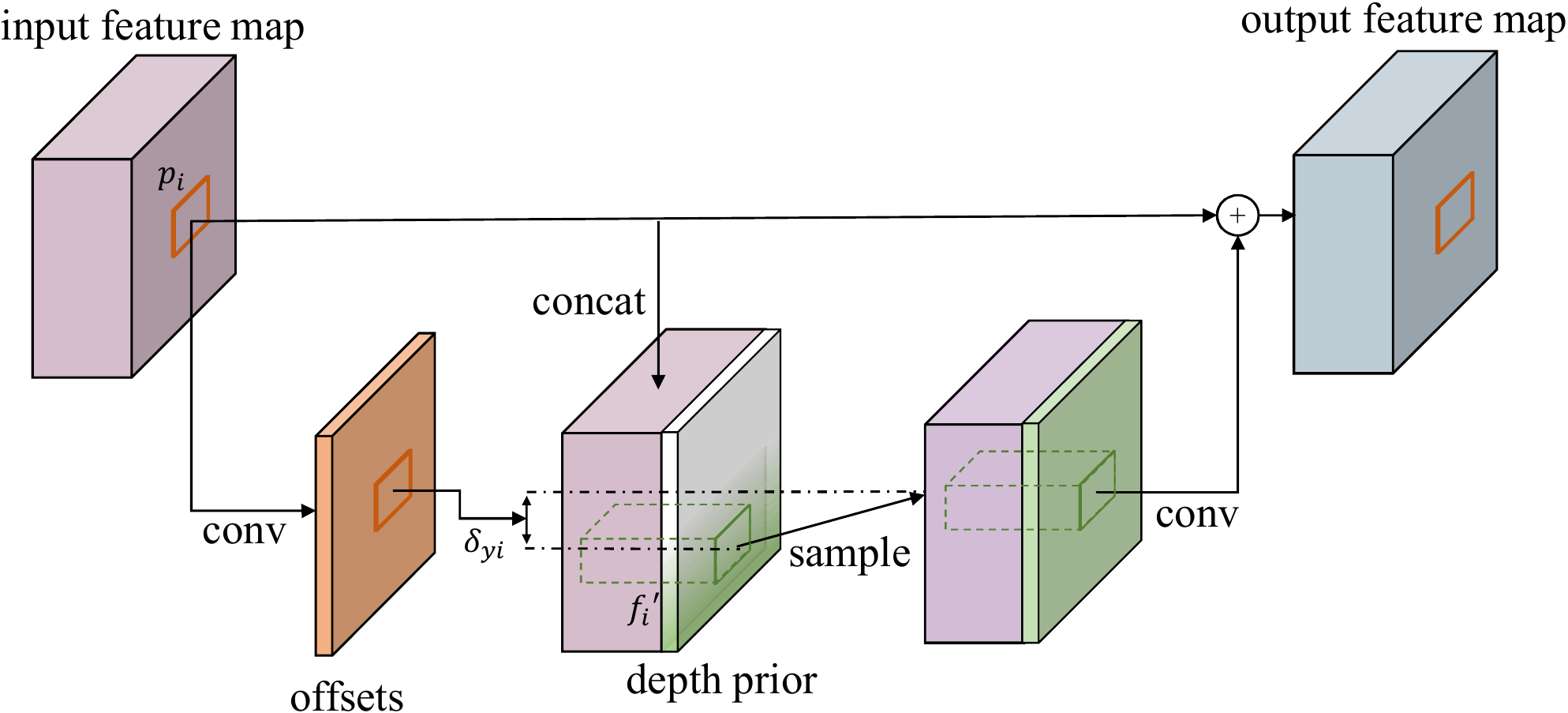}

    \caption{Ground-aware convolution. The network predicts the offsets in the vertical direction, and we sample the corresponding features and depth priors from pixels below. Depth priors are computed with perspective geometry with ground plane assumption.
    }
    \label{fig:LOOKGROUND}

\end{figure}

Ground-aware convolution is designed to guide the object center to extract features and reason the depth from its contact point; the structure is presented in Figure~\ref{fig:LOOKGROUND}.

\begin{figure}
    \centering
    \includegraphics[width=0.85\linewidth]{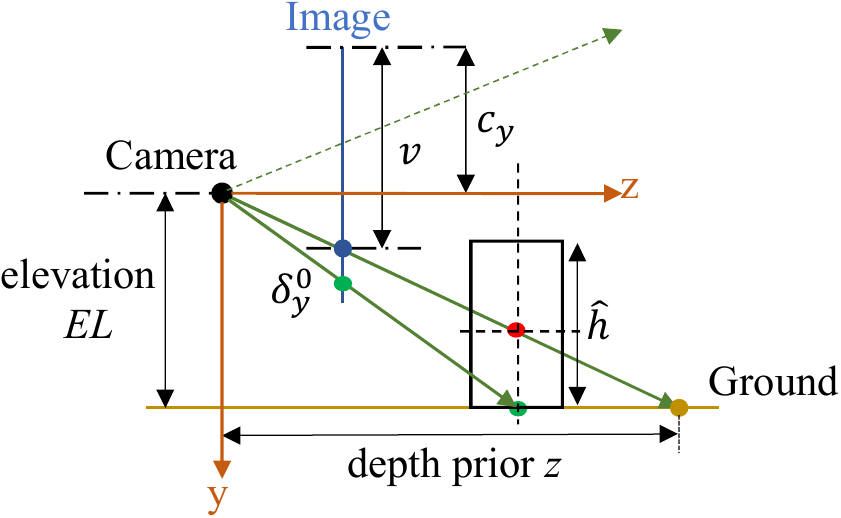}
    \caption{Perspective geometry for the GAC module. When we calculate the vertical offsets $\delta_y^0$, we assume pixels are foreground object centers. When we compute the depth priors $z$, we assume pixels are on the ground because they are features to be queried. }
    \label{fig:geometric}
  \end{figure}

To first inject perspective geometry into the network, we encode the prior depth value $z$ of each pixel point, assuming that it is on the ground. The persepctive geometry foundation is presented in Figure~\ref{fig:geometric}.

According to the ideal pin-hole camera model, the relation between the depth $z$ and height $y_{3d}$ can be obtained as:
\begin{equation}
    \label{eq:forward}
    z \cdot v = f_y \cdot y_{3d} + c_y \cdot z + T_y,
\end{equation}
where $f_y, c_y,$ and $T_y$ are focal lengths, the principal point coordinate and relative translation respectively, and $v$ is the pixel's y-coordinate in the image. 

 Assume we know the expected elevation $EL$ of the camera from the ground (1.65 meters in the KITTI dataset \cite{Geiger2012KITTI}). The distance from the ground plane pixel to the camera in $z$ can be solved from Equation \ref{eq:forward}
 as:
\begin{equation}
    \label{eq:solve_z}
    z = \frac{f_y \cdot EL + T_y}{v - c_y}.
\end{equation}

We note that the function is not continuous around the vanishing line of the ground plane ($v=c_y$), and, as indicated in Figure~\ref{fig:geometric}, physically unachievable for $v < c_y$. To detour from such a problem, we first propose to encode the depth value as the disparity of a virtual stereo setup (baseline $B=0.54 m$, similar to the KITTI stereo setup), and we derive the virtual disparity
\begin{equation}
    d = f_y \cdot B \frac{v-c_y}{f_y \cdot EL + T_y}    
\end{equation}
based on the depth $z$ in Equation \ref{eq:solve_z}. Rectified Linear Unit (ReLU) activation ($max(x, 0)$) is then applied to suppress pixels with disparity smaller than zero, which is physically unachievable for forward-facing cameras. After these two steps, the depth priors of the image becomes spatially continuous and consistent.

Inspired by CoordinateConv \cite{Liu2018CoordConv}, we treat this depth prior as an additional feature map with the same spatial size as the base feature map. Each element in the feature map is now encoded with depth priors assuming it is on the ground.

As motivated in Figure~\ref{fig:motivation}, pixels at the center of the object need to query the depth and image features from contact points, which are usually below the object centers.

Each point $p_i$ in the feature map will then dynamically predict an offset $\delta_{yi}$ as if it is the center of a foreground object
$$
\delta_{yi} = \delta^0_{yi} + \Delta_i = \frac{\hat h}{2EL - \hat h} \cdot (v - c_y) + \Delta_i,
$$, where $\hat h$ is the height of the object (we fix this to be the average height of foreground objects of the dataset), $\Delta_i$ is the residual predicted by the convolution networks. 

Then, as shown in Figure~\ref{fig:LOOKGROUND}, we extract features $f_i'$ at position $p_i + \delta_{yi}$ using linear interpolation. The extracted features $f_i'$ are merged back to the original point $p_i$ with a residual connection.

The ground-aware convolution module mimics how humans utilize the ground plane in depth perception. It extracts geometric priors and features from pixels beneath. The other part of the network is then responsible for predicting the depth residual between the priors and the targets. The module is differentiable and trained end-to-end with the entire network.

\begin{table*}[h]
    \centering
    \caption{3D Object Detection Results of Car on KITTI Test Set}
    
    \begin{tabular*}{0.8\textwidth}{ l|c|c|c|c|c|c|r}
        \toprule
        {\bf Methods} & {\bf 3D Easy} & {\bf 3D Moderate} & {\bf 3D Hard} & {\bf BEV Easy} & {\bf BEV Moderate} & {\bf BEV Hard} & {\bf Time}  \\ 
        \midrule
        MonoPSR\cite{Ku2019MonoPSR}      & 10.76 \% & 7.25 \% & 5.85 \% & 18.33 \% & 12.58 \% & 9.91 \% & 0.2s\\
        PLiDAR\cite{Weng2019Plidar}       & 10.76 \% & 7.50 \% & 6.10 \% & 21.27 \% & 13.92 \% & 11.25 \% & 0.1s\\
        SS3D\cite{Jorgensen2019SS3D}          & 10.78 \% & 7.68 \% & 6.51 \% & 16.33 \% & 11.52 \% & 9.93 \% & 0.05s\\
        MonoDIS\cite{Simonelli2019MonoDIS}      & 10.37 \% & 7.94 \% & 6.40 \% & 17.23 \% & 13.19 \% & 11.12 \% & 0.1s\\
        M3D-RPN\cite{Brazil2019M3DRPN}      & 14.76 \% & 9.71 \% & 7.42 \% & 21.02 \% & 13.67 \% & 10.42 \% & 0.16s\\
        RTM3D\cite{Li2020RTM3DRM}        & 14.41 \% & 10.34 \% & 8.77 \% & 19.17 \% & 14.20 \% & 11.99 \% & 0.05s\\
        AM3D\cite{Ma2019AM3D}         & 16.50 \% & 10.74 \% & 9.52 \% & 25.03 \% & 17.32 \% & \textbf{14.91} \% & 0.4s\\
        D4LCN\cite{Ding2019D4LCN}        & 16.65 \% & 11.72 \% & 9.51 \% & 22.51 \% & 16.02 \% & 12.55 \% & 0.2s\\
        \textbf{Ours}  & \textbf{21.65 \%} & \textbf{13.25 \%} & \textbf{9.91 \%} & \textbf{29.81 \%} & \textbf{17.98 \%} & 13.08 \% & \textbf{0.05s}\\
        \bottomrule
    \end{tabular*} \label{tab:test_results}
\end{table*}

\subsection{Network Architecture for Monocular 3D Detection}

The inference structure of the network is presented in Figure~\ref{fig:network}.
We adopt ResNet-101 \cite{He2015Resnet} as the backbone network, and we only take features at scale $1/16$.
The feature map is then fed into the classification branch and regression branch.

The classification branch consists of two convolutional layers, while the regression branch is composed of a ground-aware convolution module followed by a convolutional output layer.

The shape of the output tensor from the classification branch $C$ is $(B, \frac{W}{16}, \frac{H}{16}, K * \#anchors)$, where $K$ represents the number of classes and $\#anchors$ means the number of anchors per pixel.
    The output tensor from the regression branch is $(B, \frac{W}{16}, \frac{H}{16}, 12 * \#anchors)$.
      There are nine parameters for each anchor: four for 2D bounding box estimation, three for object center predictions, three for dimension predictions, and two more for observation angle predictions.

\subsubsection{Loss Functions}

The total loss $\mathcal{L}$ is the aggregation of classification loss for objectness $L_{cls}$, and regression loss for other parameters $L_{reg}$:
$$
    \mathcal{L} = L_{cls} + L_{reg}.
$$
We adopt focal loss \cite{Yun2018Focal}, \cite{Lin2018Focal} for classification of objectness and cross-entropy loss for the multi-bin classification
of width, height, and length.
Other parameters, $[x_{2d}, y_{2d}, w_{2d}, h_{2d}, cx, cy, z, w_{3d}, h_{3d}, l_{3d}, sin(\alpha), cos(\alpha)]$,
are normalized based on the anchors' prior parameters and optimized through smoothed-L1 loss \cite{Girshick2015Fastrcnn}.

\subsubsection{Post Optimization}
We follow \cite{Brazil2019M3DRPN} to apply hill-climbing algorithms as a post-optimization procedure.
By perturbating the observation angle and depth estimation,
the algorithm incrementally maximizes the IoU between the directly estimated 2D bounding box and the 2D bounding box projected from the 3D bounding box to the image plane.

The original implementation optimizes the depth and observation angle concurrently.
With repeated experiments, we find that optimizing only the observation angle produces even better results in the validation set. Concurrently optimizing two variables could overfit to the sparse 3D-2D constraints and affect the accuracy of the 3D prediction.

\subsection{Network Architecture for Monocular Depth Prediction}
We adopt a U-Net \cite{UNetFB15Ronneberger} structure for supervised dense depth prediction.
We select a pretrained ResNet-34 \cite{He2015Resnet} as the backbone encoder. 

In the decoding phase, the features are bilinearly upsampled, followed by two convolution layers and concat with the skip connections. We add a ground-aware convolution module before the two convolution layers in the decoder.

The depth prediction network densely predicts the logarithm of depth from each image with a $(B,1,H, W)$ tensor $y=\log{z}$. We provide supervision on each output scale $l$. The total loss is the sum of a scale-invariant (SI) loss $\mathcal{L}_{SI}$ \cite{diaz2019SoftDorn} and a smoothness loss $\mathcal{L}_{smooth}$ \cite{monodepth17} with hyperparameter $\alpha$:
\begin{equation}
    \mathcal{L} = \sum_l (\mathcal{L}_{SI} + \alpha \mathcal{L}_{smooth}).   
\end{equation}
SI loss is commonly used to simultaneously minimize the mean-square-error (MSE) and improve global consistency.
Smoothness loss is needed because the supervision from the KITTI dataset \cite{Geiger2012KITTI} is sparse and lacks local consistency.
The SI loss and smoothness loss are computed with the following equations:
\begin{align}
    \mathcal{L}_{SI} &= \frac{1}{n} \sum_i d_i^2 - \frac{\lambda}{n^2} (\sum_i d_i)^2 \\
    \mathcal{L}_{smooth} &= \frac{1}{N} \sum_i |\partial_xz^l_i|e^{-||\partial_x I^l_i||} + |\partial_yz^l_i|e^{-||\partial_y I^l_i||}, 
\end{align}
where $d_i = \log{z_i} - \log{z_i^*}$, $n$ is the number of valid pixels,  $\lambda \in [0,1]$ is a hyperparameter balancing the absolute MSE loss and relative scale loss, $N$ is the number of total pixels, and $\partial_xI$ and $\partial_yI$ are the gradients of the input images.

\section{Experiments}
\label{section:Experiments}

\subsection{Dataset and Training Setups}

We first evaluate the proposed monocular 3D detection network on the KITTI benchmark \cite{Geiger2012KITTI}. The dataset consists of 7,481 training frames and 7,518 test frames.
Chen \textit{et al.} \cite{Chen2015kittisplit} further splits the training set into 3,712 training frames and 3,769 validation frames.

We first determine the hyperparameters of the network with a family of smaller networks fine-tuned on Chen's split \cite{Chen2015kittisplit}.
Then, we retrain the final network on the entire training set with the same hyperparameters before uploading the result for testing on the KITTI server. 
The ablation study that follows is also conducted on the validation set of Chen's split.

Similar to RTM3D \cite{Li2020RTM3DRM}, we double the training set by utilizing images both from the left and right RGB cameras (only RGB images from the left camera are used in validation and final testing) and use random horizontal mirroring as data augmentation (not applied in validation and testing), which significantly enlarges the training set and improve performance. 
The top 100 pixels of each image are cropped to speed up inference, and the cropped input images are scaled to $288 \times 1280$ for the model submitted to the KITTI server, which is similar to the original scale of the images.
The feature map produced by the backbone, therefore, has a shape of $18 \times 80$. Regression loss and classification loss that are too small in magnitude (1e-3) are clipped to prevent overfitting.
The network is trained with a batch size of 8 on a single Nvidia 1080Ti GPU.
During inference, the network is fed one image at a time, and the total average processing time, including file IO and post-optimization, is 0.05s per frame.

\begin{table}
    \centering
    \caption{Depth Prediction Results on KITTI Test Set}
    
    \begin{tabular*}{0.48\textwidth}{ l|c|c|c|c }
        \toprule
        {\bf Methods} & {\bf SILog} & {\bf sqErrorRel} & {\bf absErrorRel} & {\bf iRMSE}  \\ 
        \midrule
         
        PAP\cite{PAP2019zhang}          & 13.08 & 2.72 \% & 10.27 \% & 13.95 \\
        VNL\cite{Yin2019VNLNet}         & 12.65 & \textbf{2.46} \% & 10.15 \% & 13.02  \\
        SoftDorn\cite{diaz2019SoftDorn} & 12.39 & 2.49 \% & 10.10 \% & 13.48  \\
        \midrule
        Base U-Net             & 12.78 & 3.11 \% & 10.12 \% & 13.46 \\
        \textbf{Ours}           & \textbf{12.13} & 2.61 \% & \textbf{9.41} \% & \textbf{12.65} \\
        \bottomrule
    \end{tabular*} \label{tab:depth_validation_results}
\end{table}
\begin{figure*}
    \centering
    \includegraphics[{width=1.0\textwidth}]{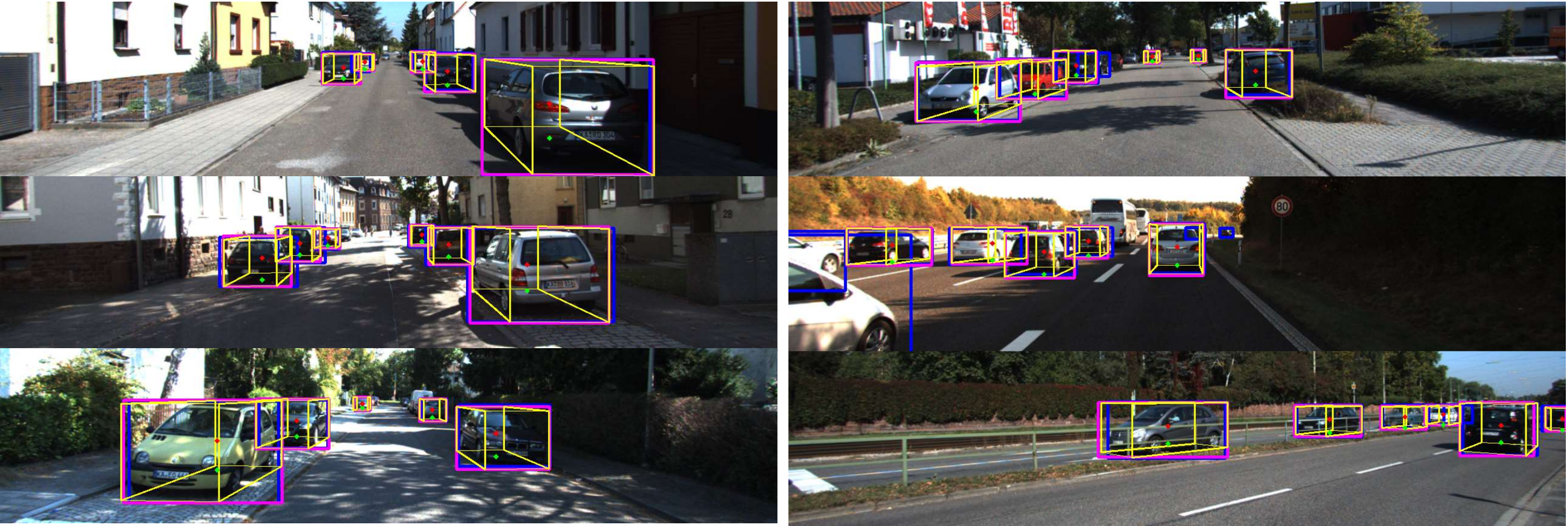}
    \caption{Qualitative examples from validation sets. Blue boxes, pink boxes and yellow boxes indicate the ground truth
    2D bounding box, estimated 2D bounding box, and estimated 3D bounding box respectively. Red points are object centers and green points visualize the offsets $\delta_{yi}$ in the GAC module.
    }

    \label{fig:examples}
\end{figure*}
\subsection{Evaluation Metric and Results for 3D Detection}

As pointed out by  Simonelli \textit{et al.} \cite{Simonelli2019MonoDIS} and the KITTI team, evaluating performance with 40 recall positions ($AP_{40}$) instead of the 11 recall positions ($AP_{11}$) proposed in the original Pascal VOC benchmark\cite{Everingham10pascal} could eliminate the problematic results presented in the lowest recall bin.
Therefore, we present our results on the test set and also ablation study based on $AP_{40}$.

The results are presented in Table~\ref{tab:test_results} alongside those of other SOTA monocular 3D detection methods based on the KITTI benchmark.

The proposed network significantly outperforms existing methods on easy and moderate vehicles. 
We do expect ground-aware convolutions to produce more accurate predictions for close-up vehicles with clear borders with the ground plane.

Qualitative results are presented in Figure~\ref{fig:examples}. The model shown here shares the same hyperparameters as the model submitted to the KITTI server but is only trained on the training sub-split. In the images on the left-hand side of the figure, cars are mostly detected and estimated accurately. The effect of the GAC module is also visualized.

We present several typical failure cases on the right-hand side of Figure~\ref{fig:examples}, and in the top-right image, the network does not detect a heavily obscured car.
In the middle-right image, truncated cars and a car that is quite far away are not detected.
We acknowledge that the network could still have trouble detecting small objects.
We show the bottom-right image to demonstrate cases in which the network give an inaccurate estimation of the 3D dimensions of a car because, as stated in Section~\ref{section:Methods}, it is still difficult to estimate the width, length, and height of an object merely by semantic information in the image. 

We provide an ablation study of the model in Section~\ref{section:Discussion}.

\subsection{Experiments on Monocular Depth Prediction}

We further evaluate the proposed depth prediction network in the KITTI depth prediction benchmark \cite{Geiger2012KITTI}.  The dataset for monocular depth prediction consists of 42949 training frames, 1000 validation samples, and 500 test samples, annotated with sparse point clouds.

The input images are cropped to $352 \times 1216$ during training and testing.  In the loss function, we applied $\alpha$= 0.3, $\lambda$=0.3 through grid-search on the validation set. The network is also trained with a batch size of 8 on a single Nvidia 1080Ti GPU. 

Scale-invariant log error (SILog) is the primary metric used in the KITTI benchmark to evaluate depth prediction algorithms. 

The results are presented in Table~\ref{tab:depth_validation_results}. The proposed network produces one of the best performances on the KITTI dataset, providing competitive results compared with SOTA methods. We also show that the network improves significantly against the baseline U-Net Model.

\begin{figure*}
    \centering
    \includegraphics[{width=0.9\textwidth}]{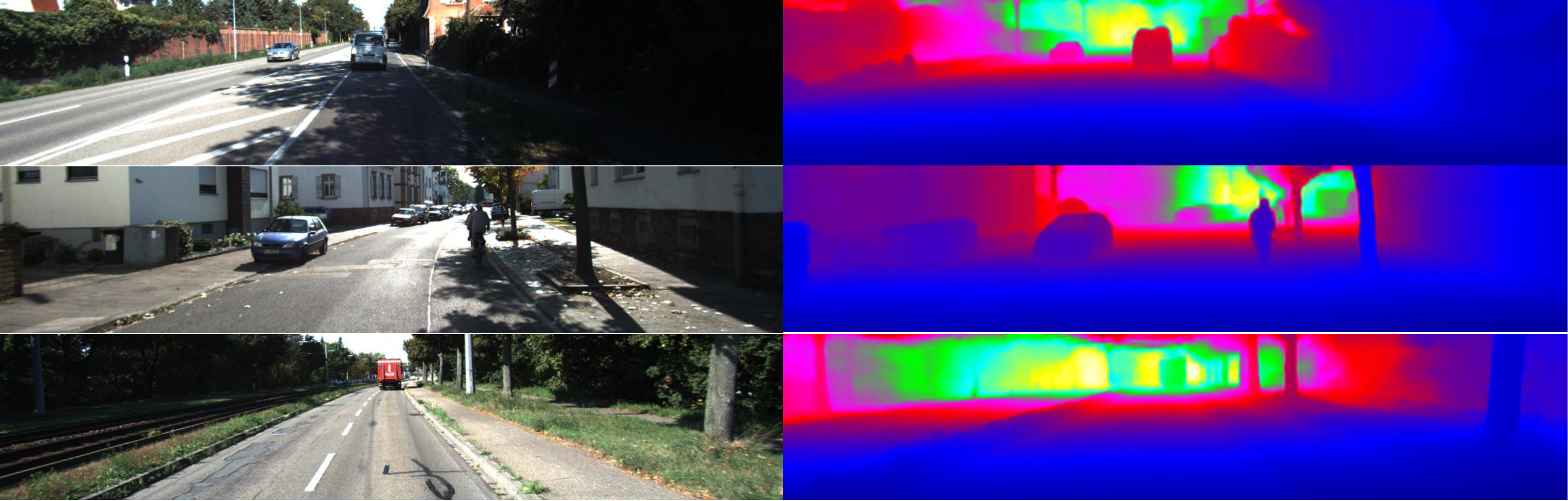}

    \caption{Qualitative examples of depth prediction from validation sets. The depth maps on the right are rendered with the official color map.
    }

    \label{fig:depth_examples}
\end{figure*}

Qualitative results are presented in Figure~\ref{fig:depth_examples}. Depth predictions inside the range of LiDAR are generally consistent. Depth predictions along long, vertical objects like trees are consistent thanks to the ground aware convolution module. We point out that there are still artifacts around the edge of objects and areas without supervision because the network receives no post-processing and little pre-training.
The depth prediction results show that the proposed module and the proposed network can improve depth inferencing from monocular images in autonomous driving scenes.

\section{Model Analysis and Discussion}
\label{section:Discussion}
In this section, we further analyze the performance of the proposed method and discuss the effectiveness of each design choice. The experiments will focus more on monocular 3D detection.
We conduct ablation studies to validate the contribution of anchor preprocessing and the ground-aware convolution module.

\begin{table*}
   \caption{3D Detection Ablation Study Results of Car on KITTI Validation Set}
   \centering
   \begin{tabular*}{0.76\textwidth}{  @{\extracolsep{\fill}} l| c|c }
      \toprule
       {\bf Methods} & {\bf $IoU\ge 0.7$ 3D Easy/Moderate/Hard } &  {\bf $IoU\ge 0.5$ 3D Easy/Moderate/Hard}  \\ \midrule
       \textbf{Baseline Model}  & \textbf{23.63 \%}/  \textbf{16.16 \%}/   \textbf{12.06} \% & \textbf{60.92 \%}/   \textbf{42.18} \%/  \textbf{32.02 \%}  \\
       \hline
    w/o Anchor Filtering & 21.39 \%/   14.35 \%/ 11.11  \% & 59.76 \%/    41.00 \%/   31.10 \% \\
    w OHEM & 22.45 \%/   15.10 \%/ 11.29  \% & 60.71 \%/    42.01 \%/   31.88 \% \\
    \hline
    w Conv                 & 21.57 \%/   15.26 \%/ 11.35  \% & 58.17 \%/    41.17 \%/   32.58 \% \\
    w DisparityConv       & 22.13 \%/   15.42 \%/ 11.34  \% & 60.13 \%/    41.62 \%/   33.07 \% \\
    w Deformable Conv      & 22.16 \%/   15.71 \%/ 11.75  \% & 62.24 \%/    43.93 \%/   33.76 \% \\

       \bottomrule
   \end{tabular*} \label{tab:ablation_study}
\end{table*}
\subsection{Anchor Preprocessing}
\label{subsection:ablation_data}
We first conduct experiments on anchor filtering. In the experiment, we do not filter out unnecessary anchors during training and testing. We notice that the proposed filtering will filter out half of the negative anchors, so we also conduct an experiment against Online Hard Example Mining (OHEM), where we filter out half of the easy negative anchors during training\cite{OHEM2016Shrivastava}.

As shown in Table~\ref{tab:ablation_study}, the baseline model outperforms the ablated one and OHEM.  The baseline model performs better at 3D inference. We also point out that there is almost no difference in 2D detection between the two models.

Generally, a one-stage single-scale object detector not only needs to classify background from the foreground but also needs to select anchors with the correct scales at foreground pixels, which also means selecting the proper depth prior. Filtering off-the-ground anchors during training and testing significantly lowers the learning burden for the classification branch of the object detector. Thus the classification branch can focus more on selecting the right anchors for foreground pixels. Such a method, as a result, also outperforms position-invariant filtering methods like OHEM.

\subsection{Ground-Aware Convolution Module}
Intuitively, basic convolutions provide a uniform receptive field for each pixel, and the network could implicitly learn to adjust its receptive field by fine-tuning the weights of multiple convolution layers. Deformable convolutions \cite{Zhu2018Deformv2} further explicitly encourage the network to adapt its receptive field according to each pixel's surrounding context. Compared with deformable convolutions, the ground-aware convolution module fixes the search direction and allows a larger search range.

We substitute the proposed module with basic convolutions, disparity-conditioned convolutions (i.e., convolution with the depth prior as an additional feature map), and deformable convolutions to examine the performance.
The results are shown in Table~\ref{tab:ablation_study}. The experiments with deformable convolutions demonstrate better 2D detection results. 

Deformable convolution can enhance the performance with a generally larger receptive field. While disparity-conditioned convolution provides the network with prior depths, the receptive field of the network is lacking. These two modules improve the performance, but the proposed module has better results by a considerable margin.

\section{Conclusion}
\label{section:Conclusion}

In this paper, we presented ground-aware monocular 3D object detection for autonomous driving scenes. 
First, we improved the problem setup for monocular 3D detection and introduced an anchor filtering procedure to inject ground plane priors and statistical priors in anchors.
Second, we introduced a ground-aware convolution module, providing sufficient hints and geometric priors for the network to reason based on ground plane priors.
The proposed monocular 3D object detection network was tested on the KITTI detection benchmark and achieved SOTA performance among monocular methods. We further tested the ground-aware convolution module in the monocular depth prediction task, and it also produced competitive results on the KITTI depth prediction benchmark.

We note that the "floor-wall" assumption is limited to scenes with specific camera poses, and it only partially holds in a complex driving scene. The proposed methods still do not reason explicitly based on the boundaries of the ground and other objects. Instead, we encode sufficient information and priors into the network and adopt a data-driven approach.

Nevertheless, the proposed method pushes the boundary of 3D detection and depth inference from images and produces powerful neural network models for autonomous driving and mobile robotics scenes.

\ifCLASSOPTIONcaptionsoff
  \newpage
\fi



%

  \bibliographystyle{unsrt}

  \bibliography{reference}

%








\end{document}